\documentclass{ecai}
\usepackage{times}
\usepackage{graphicx}
\usepackage{latexsym}
\usepackage{enumitem}
\usepackage{float}
\usepackage{graphicx}
\usepackage{multicol}
\usepackage{caption}
\usepackage{colortbl}
\usepackage{comment}
\usepackage{amssymb}
\usepackage[ruled,vlined,linesnumbered]{algorithm2e}
\usepackage[table]{xcolor}

\begin{document}

\title{Neighborhood-based Pooling for Population-level Label Distribution Learning \footnote{This is a pre-print version of a paper that was accepted for publication at European Conference on Artificial Intelligence 2020}}

\author{Tharindu Cyril Weerasooriya\footnote{Rochester Institute of Technology,
USA, email: cyriltcw@gmail.com}  \and Tong Liu\institute{Appen Machine Learning Team, USA, email: tliu@appen.com} \and Christopher M. Homan\institute{Rochester Institute of Technology,
USA, email: cmh@cs.rit.edu}}
\maketitle
\bibliographystyle{ecai}

\begin{abstract}
Supervised machine learning often requires human-annotated data. While annotator disagreement is typically interpreted as evidence of noise, population-level label distribution learning (PLDL) treats the collection of annotations for each data item as a sample of the opinions of a population of human annotators, among whom disagreement may be proper and expected, even with no noise present. From this perspective, a typical training set may contain a large number of very small-sized samples, one for each data item, none of which, by itself, is large enough to be considered representative of the underlying population's beliefs about that item. We propose an algorithmic framework and new statistical tests for PLDL that account for sampling size. We apply them to previously proposed methods for sharing labels across similar data items. We also propose new approaches for label sharing, which we call \emph{neighborhood-based pooling}.
\end{abstract}

\section{Introduction}
In supervised learning, the labels provided by a group of annotators are typically aggregated into a single label, which is regarded as ground truth. The underlying assumption that one label fits all is rarely questioned. However, the process of labeling is often subjective, i.e., based on the personal experiences of the humans who label the data \cite{ovesdotter-alm-2011-subjective}, such as when the task is to predict beauty in images or rate movies \cite{Geng2015}. Genuine disagreement is also common in seemingly ``more objective'' tasks, for instance, in assessing the mental state, beliefs, or other hidden states based on observable data \cite{Liu2019HCOMP,Homan2015}. Nevertheless, the negative impacts of AI systems trained on too narrow a segment of a population are increasingly felt \cite{Neff2016,vincent_2016,Yang_2017,pmlr-v81-buolamwini18a,Wired_2018}.

\textit{Label distribution learning} (LDL) seeks to predict, for each data item, a probability distribution over the set of labels \cite{Geng2016}. LDL can capture for each data item the diversity of opinions among the human annotators. In contrast, almost all of the prior work in LDL has taken the label distributions found in the training data to be accurate, when in fact annotations obtained from crowdsourcing sites or social media---a common though certainly not exclusive source of label distributions---are most often samples of a larger pool of annotators, who may not themselves be representative of the actual target populations' opinions. Furthermore, outside of a limited number of cases---such as movie ratings, where annotations are abundant and convenient---the sample size of annotations for each data item is far too small (often as small as five annotations per item) to reliably represent the true distributions of the annotator pool's beliefs and opinions.

\begin{figure}[h]
\centering
    \includegraphics[width=0.84\linewidth]{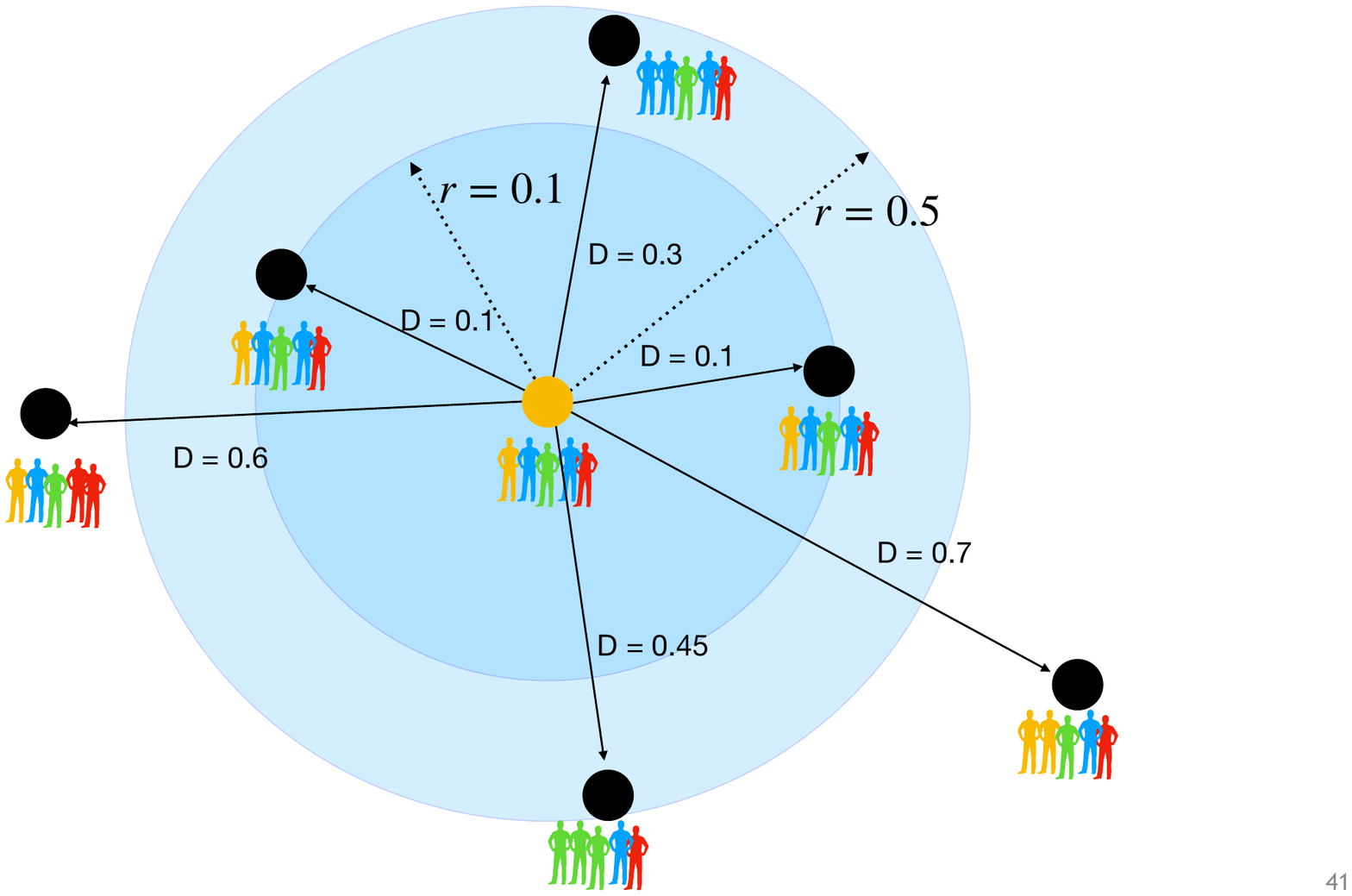}
    \caption{The neighborhood-based pooling strategy explored in this paper. The black dots represent data items. The human annotators are represented below each dot and the color of the person represents the label choice. In this example, five humans annotate each data item. $D$ represents the information theoretic divergence in each neighboring data item from the central data item and $r$ is the radius of the neighborhood. We explore the usefulness of refining the labels of the central data item for PLDL by pooling the labels of neighboring items with the central item's labels.}
    \label{fig:NBP_intro}
\end{figure}


Liu et al. \cite{Liu2019HCOMP} propose a method for sharing labels among items with similar label distributions, under the assumption that, relative to a specific labeling task, there are only a small number of possible interpretations of any data item, even at the population level. A goal in their work was to address the resource bottleneck of human annotation from a large sample by sharing the existing labels. They introduce the concept of population-level label distribution learning (PLDL), i.e., LDL that explicitly models label distributions as population samples. They explored PLDL by using different methods of clustering in the label space.  However, they do not use population models to evaluate their algorithm's performance or select models.


Our main contribution is to introduce a new model selection techniques based on population-level hypothesis tests. These techniques use traditional frequentist statistical approaches to hypothesize that items are similar if their labels can be regarded as samples from a common source. We use these methods to explore, from a sampling perspective, the clustering approach of Liu et al \cite{Liu2019HCOMP}. Additionally, we consider a new, bottom-up approach, called \emph{neighborhood-based pooling} (NBP) (Figure \ref{fig:NBP_intro}), for improving the ground truth estimates of labels for PLDL. Our approach is based on the idea that items with very similar labels may have essentially the same meaning, relative to an annotation task, and thus may be shared, but that the space of label distribution is less clustered and smoother, rather than clustered, as Liu et al. assume. 


 
\section{Related Work}
Disagreement during labeling tasks is a well studied problem \cite{Malossini2006,Marcus1993,ovesdotter-alm-2011-subjective}. Snow et al. \cite{Snow2008} observed using multiple crowdsourced workers that individuals (even experts) had personal biases when labeling. This can contribute to diversity among labels through disagreement, which nondistributional learning cannot account for. Aroyo \cite{Aroyo2014} observed that when the human annotators agree with one another, they perform at a level comparable to experts, and when they disagree, it is often for a good reason. In conventional single label learning applications, researchers consider the majority label as the ground truth, disregarding previously discussed disagreements. 

Geng \cite{Geng2016} introduced label distribution learning, a learning paradigm where probability distributions are objects to be predicted. LDL takes into account the diversity, disagreement, ambiguity, and uncertainty between annotators for its approaches including, predictions. He and colleagues studied LDL using a variety of different learning algorithms, problem transformation, algorithm adaptation, and a specialized algorithm. The algorithm for LDL shares some similarities with \textit{learning over a probability distribution}, which has a long history of research \cite{Sheng2008,Goyal2018,Algorithm1979}. Both use probability distributions; LDL interprets them as ground truth while the others use them to model uncertainty. Geng and other researchers studied the applications of LDL in various settings such as predicting population level labels \cite{Geng2015,Geng2014,Ren2017} while some do not \cite{Gao2017,Ling2018}. These studies acknowledge the need for a large number of labels to train on for improving their distributions. 
We use the natural scene and facial expressions datasets used by Geng \cite{Geng2014} for our experiments (see Section \ref{experiments}).  



In contrast to supervised learning, semi-supervised learning (SSL) is a combination of supervised and unsupervised learning \cite{Chapelle2006}. SSL uses both labeled and unlabeled data to improve learning. SSL \cite{Iscen2019,zhu05survey} has a long history with a variety of methods of learning. Clustering is another semi-supervised learning method in which if two items belong to the same cluster, they are believed to share the same label \cite{Chapelle2006}. 

Liu et al. \cite{Liu2019HCOMP} use a similar approach for PLDL (they also formally introduce the PLDL problem). They explored clustering as an unsupervised learning method to improve the quality of ground truth estimation. They \cite{Liu2019HCOMP} collected a crowdsourced dataset in which five-to-ten human annotators labeled each data item. Then they ``[clustered] together semantic similar data items, and then [pooled] together all the labels in each cluster into a single, larger sample'' \cite{Liu2019HCOMP}, under the assumption that this sample represented the population-level beliefs about each item in the cluster. They compared and contrasted the performance of a variety of clustering methods. Their approach is more typically applied to items with no or unreliable labels, where similarity between data items is necessarily determined in the feature space of the data items, not in the label space. However, Liu et al. \cite{Liu2019HCOMP} observe that, if all the items already have labels, one can determine similarity in the label space alone.  Their work has two limitations: (1) though their approach was based on statistical principles, they do not exploit this connection in their analysis and (2) they consider only a relatively small number of clusters per learning problem (under the assumption that for classification problems the number of distinct answer distributions is necessarily limited).

Zhang \cite{Min-LingZhang2005} introduces multi-instance-multi-learning (MIML)-$k$-NN as a non-parametric learning method that used the \textit{k}-nearest neighbor for multiple labels. However, $k$-NN selects the closest $k$ neighbors around the item regardless of how close or further away they are from the data point. The drawback of this approach is not taking into account the similarity between the item and its neighbors.  

\section{Methods}
\subsection{Label Distribution Learning}

Wang and Geng \cite{wang2019theoretical} developed a theory for label distribution learning, proving a number of theorems about error functions for LDL. Their theory presents label distributions as ground truth objects, without considering that they may be merely estimates of an underlying ground distribution. Here, we present a theory for LDL for the special case of when the label distribution is a sample of an underlying population, which explicitly accounts for this population, along with noise in the sampling process and level of reliability of the annotators. First, we introduce some notation. For any probability distribution $\mathcal{D}$, and any set $\mathcal{X}$ let $X_\mathcal{D}$ denote a random variable in $\mathcal{X}$ over $\mathcal{D}$. If context is clear we will drop the ``$\mathcal{D}$'' subscript. Let $\mathcal{D} | X$ denote the distribution conditioned on $X$, and similarly for $x \in \mathcal{X}$ and $\mathcal{D} | x$. Thus, $Y_{\mathcal{D} | x}$ is a random variable in $\mathcal{Y}$ over the distribution $\mathcal{D} | x$. Finally, for any set $\mathcal{X}$, let $\mathcal{P}_{\mathcal{X}}$ denote the space of all probability distributions over $\mathcal{X}$.

\begin{algorithm}
$S \leftarrow ()$\\
\For{$i\leftarrow 1$ \KwTo $n$}{
choose $x_i \sim X_\mathcal{D}$\\
\For{$j \leftarrow 1$ \KwTo $m(x_i)$}{
    choose $a_{i,j} \sim P_\mathcal{D}$\\
    choose $y_{i,j} \sim Y_\mathcal{D}~|~x_i,a_{i,j}$\\
    add $(x_i, a_{i,j}, y_{i,j})$ to $S$
}
}
\Return{S}
\caption{Sampling procedure.}
\label{alg:pldl_sample}
\end{algorithm}

Now, to formally define PLDL, let  $\mathcal{D}$ be a probability distribution over $\mathcal{X} \times \mathcal{A} \times \mathcal{Y}$, where $\mathcal{X}$ is the \emph{feature space} of a data set of interest, $\mathcal{A}$ is the \emph{agent (or annotator) space}, and $\mathcal{Y}$ is a \emph{label space}. Let $X_{\mathcal{D}}$, $A_{\mathcal{D}}$, and $Y_{\mathcal{D}}$ be random variables representing the marginal distributions of $\mathcal{X}$, $\mathcal{A}$, and $\mathcal{Y}$ respectively. We assume that $X_{\mathcal{D}}$ is independent from $A_{\mathcal{D}}$. 


Let $\mathcal{S}$ be a sample from $\mathcal{D}$, drawn according to Figure \ref{alg:pldl_sample}. We can thus consider the indexed set $(x_i)_{p(X_\mathcal{S} = x_i) > 0}$.
and let $\mathcal{H}$ be a \emph{hypothesis space}, where each $h \in \mathcal{H}$ is a function $h:\mathcal{X} \rightarrow \mathcal{P}_{\mathcal{Y}}$.
The \emph{empirical risk minimization (ERM) problem} for PLDL is to find a hypothesis $h_\mathcal{S} \in \mathcal{H}$ that minimizes $L_\mathcal{S}$, the \emph{loss function} applied to $\mathcal{S}$ on $h$:
\begin{eqnarray}
h_\mathcal{S} &\in& \arg\min_{h \in \mathcal{H}} L_\mathcal{S}(h)
\end{eqnarray}
(Digression: in the multi-label setting one can take $\mathcal{Y'} = 2^{\mathcal{Y}}$ and treat it as a single label learning problem; however, it is often beneficial to exploit the set structure of the multi-label setting in designing a machine learning solution.)
$L_\mathcal{S}$ is a function that is small when each $h(x_i)$ is close to $Y_{\mathcal{S}|x_i}$ and zero whenever $h(x_i) = Y_{\mathcal{S}|x_i}$. For the sake of discussion, we will take $L_\mathcal{S}$ to be the expected Kullback-Liebler (KL) divergence,
$E_\mathcal{S}[\mathrm{KL}(h(x_i)||Y_{\mathcal{S}|x_i})]$, where for any two probability distributions $\mathcal{P}$ and $\mathcal{Q}$ with random variables over $\mathcal{Y}$,

\begin{eqnarray}
\mathrm{KL}(P||Q) &=& \sum_{y \in \mathcal{Y}} P(y)\log\left(\frac{P(y)}{Q(y)}\right)
\end{eqnarray}
KL divergence is widely used in machine learning, especially in belief modeling, and as a loss function in many settings. Here, it can be roughly interpreted as the expected number of bits per item needed to correct whenever a sample from $P$ is mistaken for a sample from $Q$, and this seems to capture intuitively the notion of error when comparing two probability distributions.

\subsection{Estimating Ground Truth via Label Pooling}
A common problem that occurs in population-based label distribution is that datasets frequently only have a small number of annotations per data item, i.e., each $x$ occurs at most $m$ times in $\mathcal{S}$, where $m$ is typically too small to estimate the ground truth sample $f(x)$. For simplicity's sake we will assume hereafter that each item occurs exactly $m$ times. 

Liu et al. \cite{Liu2019HCOMP} explore the idea that similarly-labeled data items may have similar meanings at the population level and could thus be seen as samples of a common underlying source. We generalize the idea of pooling and extend it to neighborhood-based approaches, but first we need to formally define pooling: A \emph{pooling} is: an integer $p \in \mathbb{N}$, a collection of sets $K_1, \ldots, K_p \subseteq \mathcal{X}$ such that $K_1 \cup \cdots \cup K_p = \mathcal{X}$, and a mapping $k: \mathcal{X} \rightarrow \{1, \ldots p\}$. After learning a model $(p, \{K_1, \ldots, K_p\}, k)$ that best fits the data, each data item $x \in \mathcal{X}$ is then associated with the marginal label distribution $Y_{K_{k(x)}}$ (which Liu et al. call the \emph{refined label distribution}) of $x$'s cluster $K_{k(x)}$.


\subsubsection{Cluster-Based Pooling}
Liu et al. \cite{Liu2019HCOMP} introduce pooling methods based on generative hierarchical probabilistic models. Each of these models assumes that the empirical label distributions were generated by choosing a cluster according to a distribution $\pi$ (or in the case of latent Dirichlet allocation (LDA) \cite{blei2003latent}, each data item $x_i$ has its own distribution $\pi_i$) over the pools, and then choosing the labels via a distribution $\phi_j$ associated with the chosen cluster (or in the case of LDA, choosing a new ``pool'' for each label). 

For comparison purposes, we consider four of the models used by Liu et al. \cite{Liu2019HCOMP}: a (finite) multinomial mixture model (\textbf{F}) with a Dirichlet prior over $\pi \sim \mbox{Dir}(p, \gamma = 75)$, where each cluster distribution $\pi_j$ is a multinomial distribution with Dirichlet priors $\mbox{Dir}(d, \gamma = 0.1)$, a Gaussian mixture model 
(\textbf{G}) without Dirichlet priors, k-means (\textbf{K}) and LDA (\textbf{L}).

\subsubsection{Neighborhood-based Pooling}
Neighborhood based pooling (NBP) creates for each data item $x \in \mathcal{S}$ one pool $K_x = \{x'~|~ D_{KL}(Y_{\mathcal{S}|x}||Y_{\mathcal{S}|x'}) < r\}$,
where $r > 0$ is a hyperparameter called the \emph{neighborhood radius} (see Algorithm \ref{alg:nbp_generator}). 

\begin{figure}[h]
  \centering
  \includegraphics[width=\linewidth]{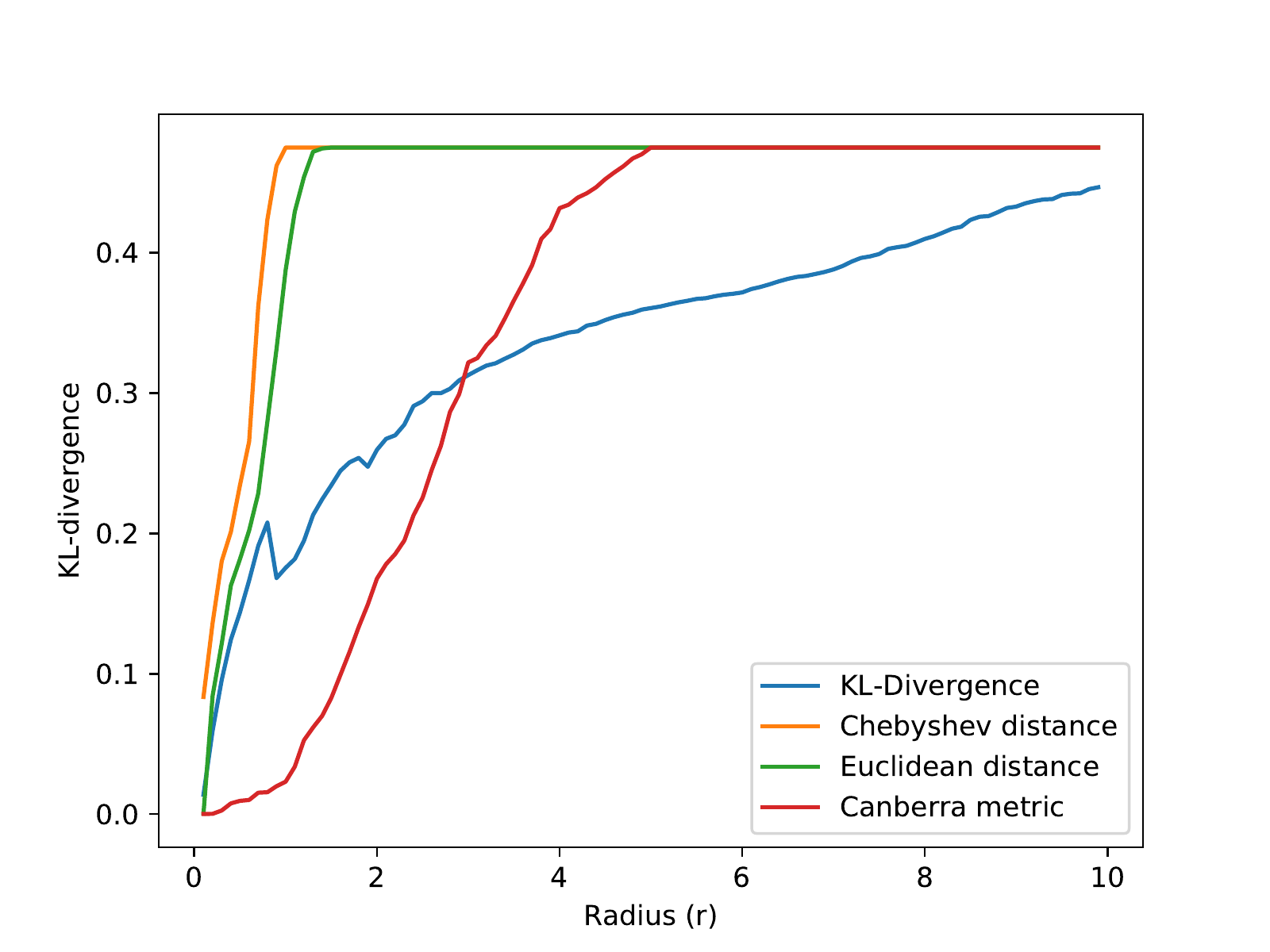}
  \caption{The KL-divergence for each $r$ value for the \emph{JQ1} dataset with various distance/divergence measures.}
 \label{fig:nbp_distance}
\end{figure}

We also considered Euclidean distance, Chebyshev distance, and the Canberra metric, however our results using these methods were similar enough to our results with KL divergence (Figure~\ref{fig:nbp_distance}). Additionally, KL divergence has a meaningful interpretation in the context of statistical estimation. It is the expected number of bits per item needed to make a multinomial sample from one distribution appear to be from the other.

\begin{algorithm}[h]
\SetAlgoLined
\smaller
\caption{Neighborhood based pooling (NBP)}
\label{alg:nbp_generator}
\SetKwProg{generate}{Function }{}{end}
\textit{\textbf{Inputs:}}\\
Empirical label distributions $(Y_{\mathcal{S}|x_i})_{i \in \{1,\ldots, n\}}$\\
Radius of the neighborhood $r$ \\
Information theoretic measure $D$\\
\textit{\textbf{Output:}}\\
A pooling $(p, (K_1, \ldots, K_n), k)$\\
\generate{Neighborhood Based Pooling}{
     \ForAll{$i \in \{1,\ldots,n\}$}{
     let $K_i = \{x \in \mathcal{X}_\mathcal{S} ~|~ D(Y_{\mathcal{S}|x} || Y_{\mathcal{S}|x_i}) \leq r\}$\\
     let $k(x_i) = i$
    }
    return $(p, (K_1, \ldots, K_n), k)$
}
\end{algorithm}

\subsection{Hyperparameter Selection for (and Evaluation of) Pooling Models}
\label{ss:pooleval}
To evaluate our label pooling methods and select hyperparameters (the number of pools $p$ for the clustering or the neighborhood radius $r$ for the NBP methods), we consider two loss functions: first, we use the mean KL divergence between the empirical and label distributions of each item the evaluation set.

\begin{eqnarray}
\mathcal{L}_S(p, (K_1, \ldots, K_p),k) = \sum_{i \in \{1,\ldots,n\}} D_{KL}(Y_{K_{k(x_i)}} || Y_{\mathcal{S}|x_i})/n
\label{eqn:raw_loss}
\end{eqnarray}

Second, we consider the probability of the given label set, given the pools. Additionally, the clustering methods all rely on stochastic optimization. To avoid overfitting, for a range of cluster sizes $p \in \{1,\ldots 40\}$ (based on the range selected by Liu et al. \cite{Liu2019HCOMP}), we ran 100 trials on the training set (since we are using clustering here for sample estimation and not prediction, it is valid and proper to test on the training set) picked the model for each value of $p$ with the median loss. Table~\ref{table:model_selection_cluster_kl} shows the number of clusters selected on each label set. 

Selecting the best hyperparameter by the raw loss function (Eq \ref{eqn:raw_loss})  may not be the best choice here, in part because it only measures which hyperparameter fits the model best, but also because ground truth is for us unobservable. Sometimes, even the best models are still not adequate. In unsupervised problems such as pooling, there are few settings where widely agreed-upon numerical measures are sufficient to judge the quality of a model \cite{Halkidi2001}. 

However, in the case of PLDL we have a population of annotators $\mathcal{A}$ to work with. This enables us to use frequentist statistical methods such as hypothesis testing to assess the quality of our pooling models. The basic idea is to generate from each given model random samples that are the same size as our training sample. If the model is a good fit for our training sample, then statistics from the training sample should be in line with those generated directly by the model. 

So for each model, we run two simulation-based statistical tests. First we generate 1000 synthetic label sets, each the size of our training set, based on the pooling model we are testing, we then compare the loss function on our test data to the distribution of loss functions on the synthetic sets. As an additional test, we compare these to another sample of 1000 synthetic label sets based on a bootstrap sample from the actual data. See Algorithms \ref{alg:model_selection}--\ref{alg:model_selection_cluster}.

\begin{eqnarray*}
\lefteqn{L_\mathcal{S}(p, (K_1, \ldots, K_p), k)}\\
&=&\log n! + \sum_{i=1}^n \sum_{y \in \mathcal{Y}}y_{\mathcal{S}|x_i}^{\#} \log y_{K_{k(x_i)}} - \log y_{\mathcal{S}|x_i}^{\#}!,
\end{eqnarray*}
where $y_{\mathcal{S}|x_i}^{\#}$ denotes $|\{y \in \mathcal{Y}~|~\exists x, a((x,a,y) \in \mathcal{S}\}|$.

We consider a range of reasonable values for $r$ based on our NBP and bootstrap sampler. Table~\ref{table:nbp_model_selection} shows the neighborhood sizes ($r$) selected on each of the label sets. The selected $r$ is the elbow point of a piecewise linear regression line (See Figure~\ref{figure:nbp_jobq1_diff_boot} and Figure~\ref{figure:nbp_jobq1_diff_nbp}). 

\begin{algorithm}[h]
\SetAlgoLined
\smaller
\caption{Model Selection/Evaluation for Pooling}
\label{alg:model_selection}
\SetKwProg{generate}{Function }{}{end}
\textit{\textbf{Inputs:}}
 \quad A sample of data items and empirical label distributions $\mathcal{S}$ \\
 \quad A pooling of the data $(p, (K_1, \ldots, K_p), k)$\\
 \quad A stochastic sample generator $G$ for random label samples\\ 
 \quad A statistic $L$ defined over labeled data items and poolings of data\\
 \quad The number votes $m$ per item.\\
 \quad The number $b$ of samples used for testing\\
\textit{\textbf{Output:}}
\quad The fraction of samples $\mathcal{B}$ such that $L_{\mathcal{B}} (p, (K_1, \ldots, K_p), k) > L_\mathcal{S}(p, (K_1, \ldots, K_p), k)$\\
\generate{Model Evaluation}{
$count \leftarrow 0$\\
     \For{$i \in \{1,\ldots, b\}$}{
     generate $\mathcal{B} \sim G((p, (K_1, \ldots, K_p), k), n, m)$ according to \\
     ~~~~~~one of the sampling procedures from Algorithm \ref{alg:pldl_sample}\\
     \If{$L_\mathcal{B}(p, (K_1, \ldots, K_p), k) > L_\mathcal{S}(p,(K_1, \ldots, K_p), k)$}{$count \leftarrow count + 1$}
     }
    return $count / n$ 
}
\end{algorithm} 

\begin{algorithm}[h]
\SetAlgoLined
\smaller
\caption{Sampling approaches}
\label{alg:model_selection_cluster}
     $G = $ cluster sampler\\
     \For{$i \in \{1,\ldots, n\}$}{
     choose a cluster $K_j \sim \pi$\\
     choose $Y_{\mathcal{B}|x_i} \sim \mathcal{M}(Y_{K_j}, m)$ where $\mathcal{M}(Y_{K_j}, m)$ denotes the multinomial distribution over all size $m$ i.i.d. samples of $Y_{K_j}$\\
     }
    \hrulefill\\
     $G = $ NBP sampler\\
    \For{$i \in \{1,\ldots, n\}$}{
     choose $j \in \{1,\ldots n\}$ uniformly at random\\
      choose $Y_{\mathcal{B}|x_i} \sim \mathcal{M}(Y_{K_j}, m)$\\
    
     }
     \hrulefill\\
      $G = $ Bootstrap sampler\\
     \For{$i \in \{1,\ldots, n\}$}{
     choose $j \in \{1,\ldots n\}$ uniformly at random\\
     choose  $Y_{\mathcal{B}|x_i} \sim \mathcal{M}(Y_{\mathcal{S}|x_j}, m)$
     }
\end{algorithm}

\section{Experiments} \label{experiments}
\subsection{Data and Labels}
All of the datasets we consider have labels that were produced by humans, and in many cases so was the data, we consulted with our institutional review board (IRB), who determined that the data was both secondary and publicly available, and thus exempt from IRB review. For each dataset, we used a 50/25/25 train/dev/test split. These datasets have been used in prior LDL related research \cite{Liu2019HCOMP,Geng2015}. 

\textbf{Jobs dataset - JQ1, JQ2, and JQ3} We used a set of 2,000 job-related annotated tweets data set collected by Liu et al. \cite{P16-1099} for this research. The dataset contains responses from three annotation tasks. Originally five crowdsourced annotators each from Mechanical Turk and FigureEight (10 annotators total) labelled the data. The task for JQ1 was to identify the point of view of the tweet (i.e., 1st person, 2nd person, 3rd person, unclear, or not job related). JQ2 was to capture the employment status of the subject in the tweet (i.e., employed, not in labor force, not employed, unclear, and not job-related). The final task was to identify if there was any mention of a employment transition event in the tweet (i.e., getting hired/job seeking, getting fired, quitting a job, losing job some other way, getting promoted/raised, getting cut in hours, complaining about work, offering support, going to work, coming home from work, none of the above but job related, and not job-related).

\textbf{Natural Scenes (NS) dataset\footnote{\label{footnote1}The datasets are to download on the website http://ldl.herokuapp.com/download}}
We also use the natural scenes dataset by \cite{Geng2016}. This set contains 2,000 images of natural scenes (NS). Each image has a label distribution over nine labels (i.e., plant, sky, cloud, snow, building, desert, mountain, water, and sun) and with labels collected from ten human annotators. These images  have 36 features associated with them. 

\textbf{BU-3DFE Facial Expression (FE) dataset}
This dataset contains 2,500 facial expression images, each of these images are associated with a label distribution over 6 label categories (i.e., happiness, sadness, surprise, fear, anger, and disgust). The dataset was collected by Yin et al \cite{face_dataset}. For each image, the labels come from 23 human annotators.  

\subsection{Experiments on Label Pooling}
\begin{figure*}
    \includegraphics[width=\textwidth]{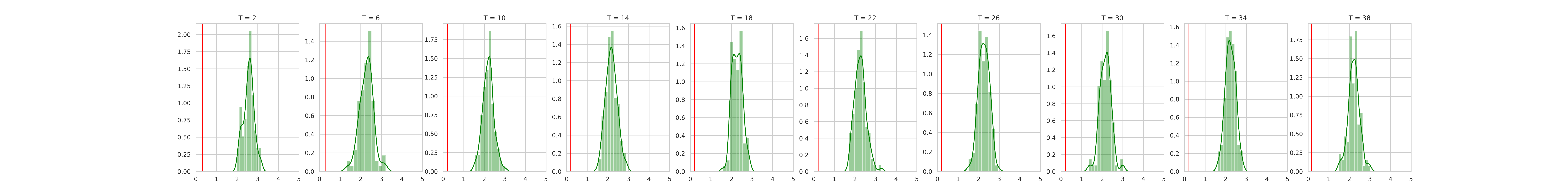}
    \includegraphics[width=\textwidth]{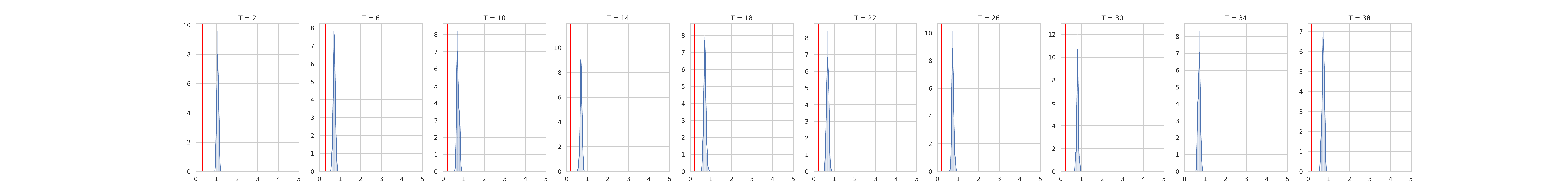} 
  \caption{Comparison of the $\mathcal{L}_{\mathcal{B}}(p, (K_1,\ldots, K_p), k)$ (synthetic dataset) distributions obtained from the sampling techniques and clustering methods on the \textit{JQ1} dataset for LDA. Reference line: $\mathcal{L}_{\mathcal{S}}(p, (K_1,\ldots, K_p), k)$ value of the training dataset. Top: \textit{Cluster sampler in green.} Bottom: \textit{Bootstrap sampler in blue.}\label{figure:lda_jobq1_histogram}}
  \begin{minipage}{0.48\linewidth}
    \includegraphics[width=\linewidth]{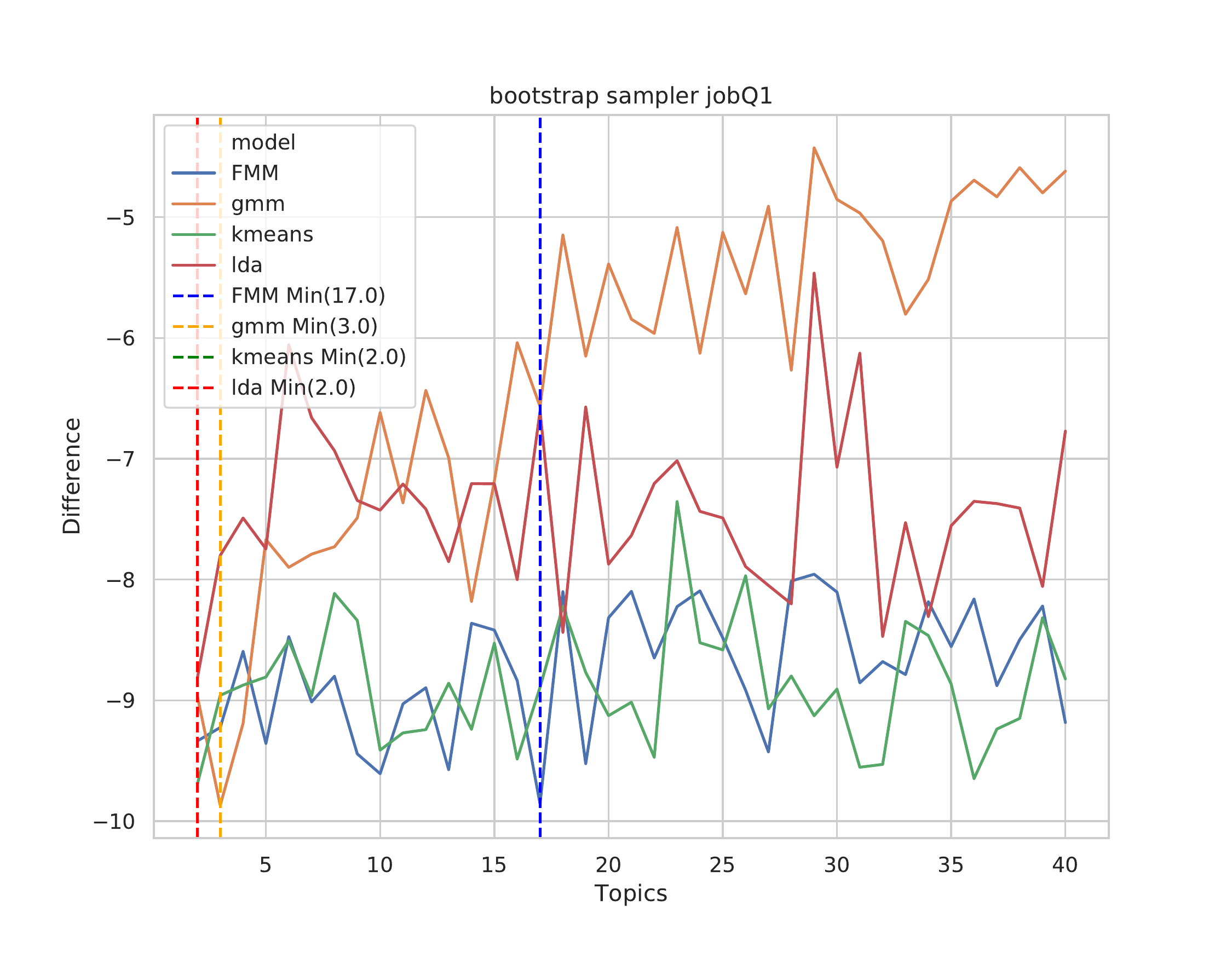}
     \caption{The difference between the average KL-divergence of the sample generated by the bootstrap sampler and the predicted data for the JQ1 dataset with the bootstrap sampler. \label{figure:lda_jobq1_diff_boot}}
    
\end{minipage}%
    \hfill%
\begin{minipage}{0.48\linewidth}
  \includegraphics[width=\linewidth]{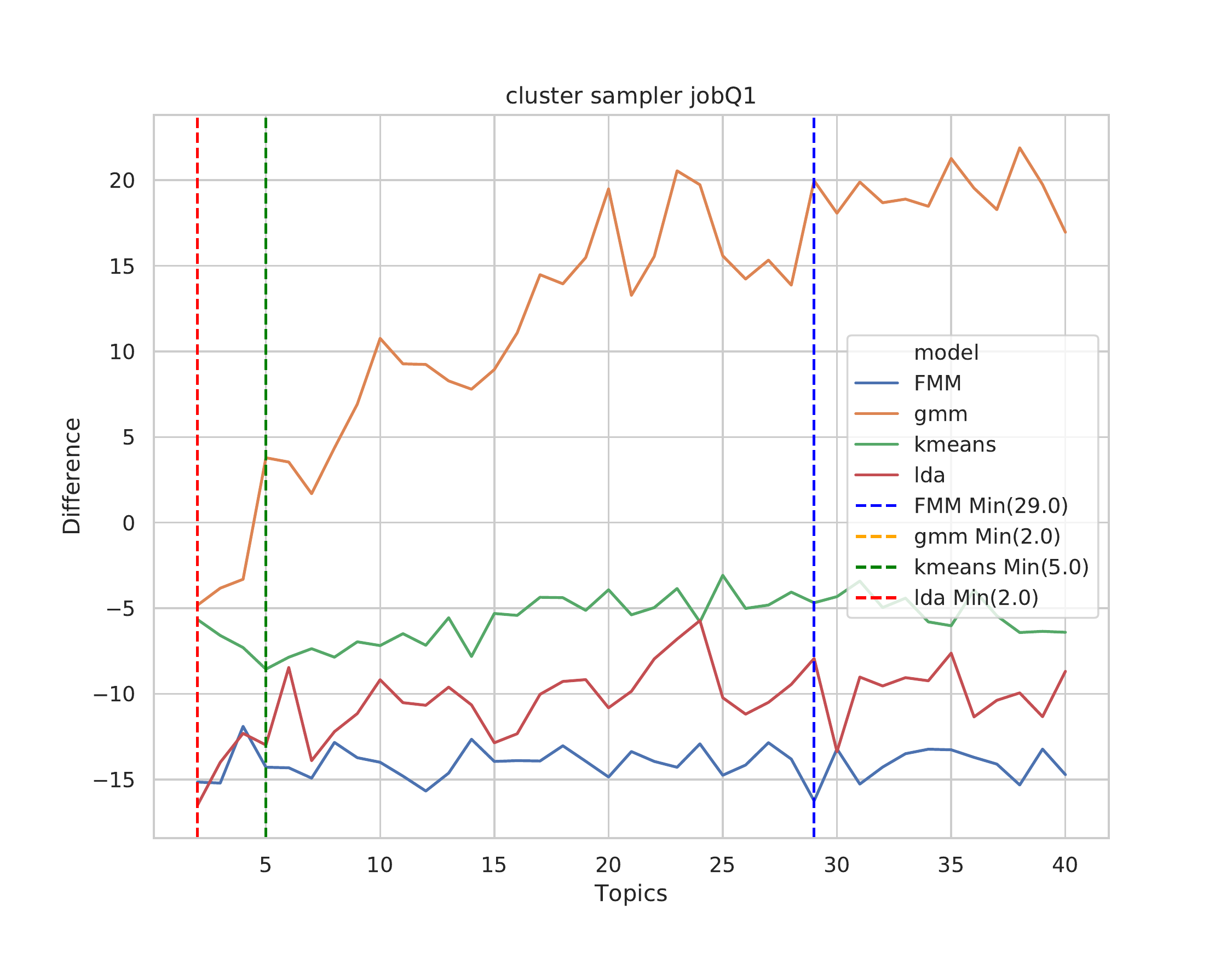}
       \caption{The difference between the average KL-divergence of the sample generated by the cluster sampler and the predicted data JQ1 dataset with the cluster sampler.}
       \label{figure:lda_jobq1_diff_cluster}
\end{minipage} 
\begin{minipage}{0.48\linewidth}
    \includegraphics[width=\linewidth]{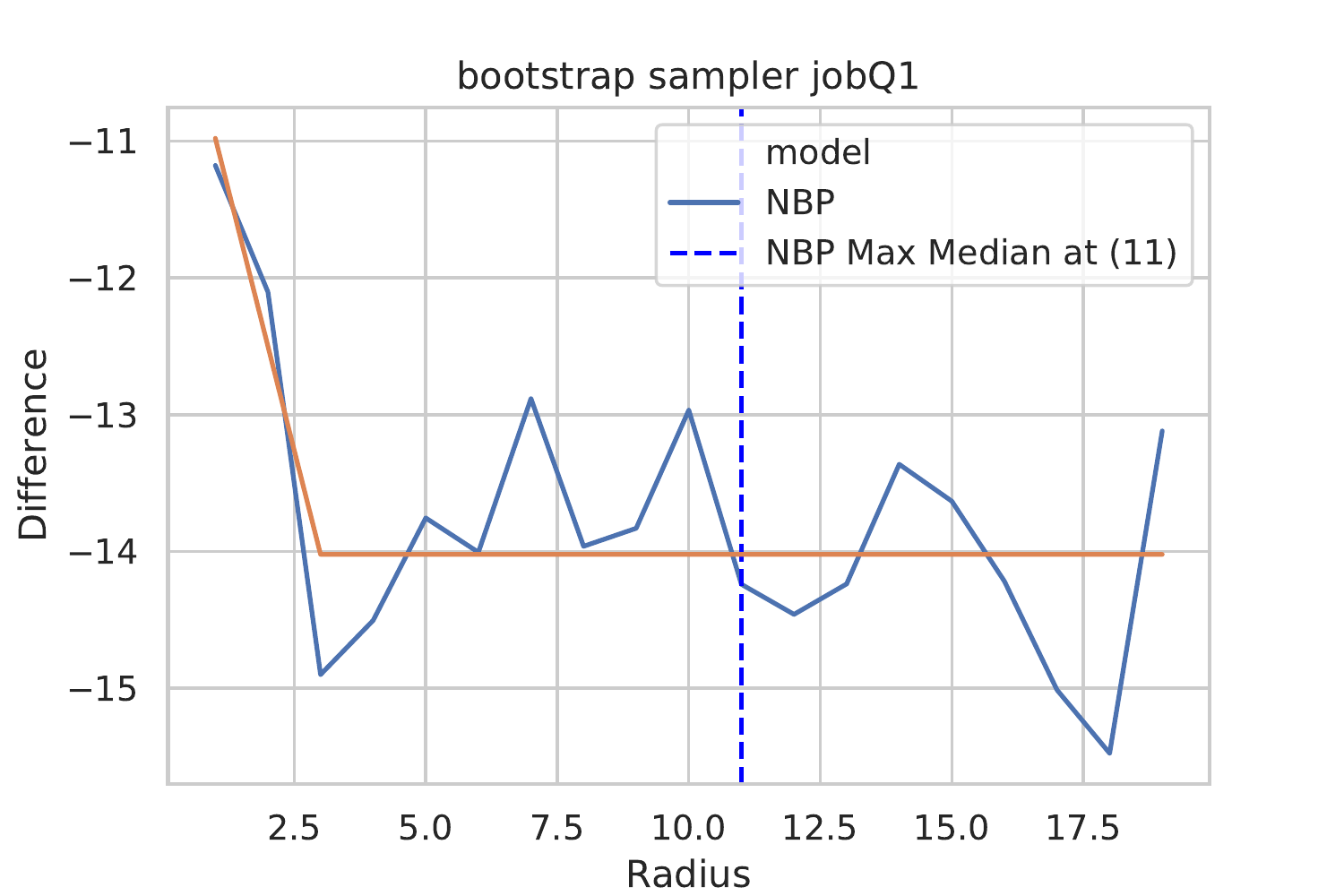}
    \caption{The difference between the average KL-divergence of the sample generated by the bootstrap sampler and the predicted data for the JQ1 dataset with the bootstrap sampler for NBP. \textit{We included for reference the piece-wise linear regression line.}\label{figure:nbp_jobq1_diff_boot}}
\end{minipage}%
    \hfill%
\begin{minipage}{0.48\linewidth}
  \includegraphics[width=\linewidth]{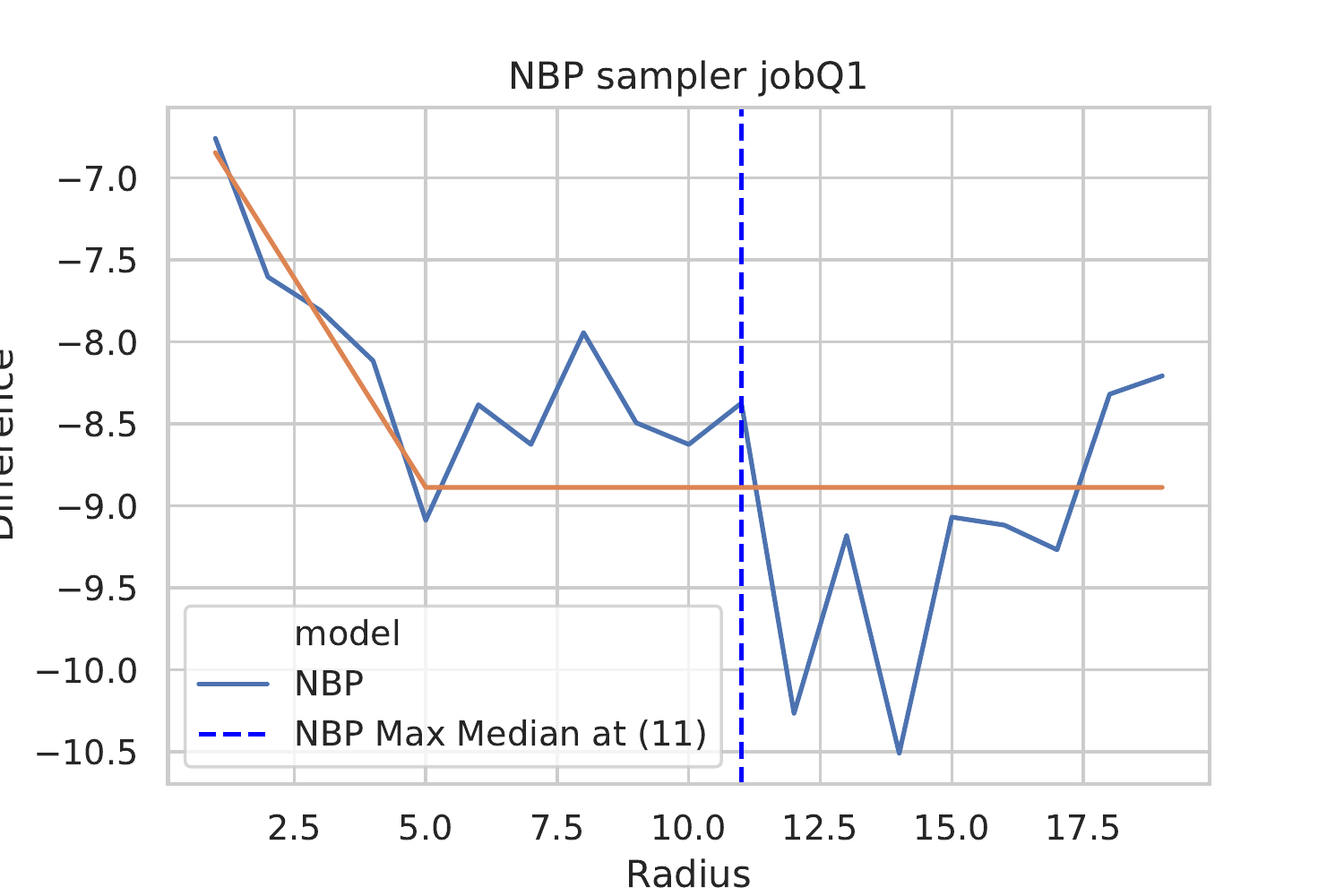}

\caption{The variation of the standard difference between the average KL-divergence of the sample generated by the NBP sampler and the predicted data JQ1 dataset with the NBP sampler. \textit{We included for reference the piecewise linear regression line.}\label{figure:nbp_jobq1_diff_nbp}}
\end{minipage} 
    \hfill%
\end{figure*}

In evaluating our pooling results according to the simulation-based methods described in Section \ref{ss:pooleval} we noticed that, frequently, the values of the loss function on the synthetic data were either all greater or all less than their corresponding values in the training data (recall that we evaluated the model on the training data because, given its purpose to estimate population statistics from samples, rather than predict individual values on new items, held-out data was not necessary). This made our test defined by Algorithm \ref{alg:model_selection} meaningless. Thus, rather than use it, we instead, for each parameter value $\phi$ to be tested, and corresponding pooling $(p, (K_1,\ldots, K_p), k)$ subtract the loss on the training data $\mathcal{L}_{\mathcal{S}}(p, (K_1,\ldots, K_p), k)$ from the mean loss on the synthetic data $\mathcal{L}_{\mathcal{B}}(p, (K_1,\ldots, K_p), k)$, divided by the standard deviation of the synthetic data loss:
\begin{eqnarray}
\frac{\mu(\mathcal{L}_{\mathcal{B}}(p, (K_1,\ldots, K_p), k)) - \mathcal{L}_{\mathcal{S}}(p, (K_1,\ldots, K_p), k)}{\sigma(\mathcal{L}_{\mathcal{B}}(p, (K_1,\ldots, K_p), k))}
\end{eqnarray}
and seek the parameter $\phi$ that minimizes this quantity. The optimal number of clusters ($p$) based on the observation for $\phi$ is given in Table~\ref{table:model_selection_cluster_md} and Table~\ref{table:model_selection_cluster_kl}. For the JQ1, JQ2, and JQ3 datasets, \textbf{F} clustering models outperformed other clustering methods based on these results. It was followed by \textbf{L} clustering for the same set of labels. As expected, \textbf{K} and \textbf{G} come last. Figure~\ref{figure:lda_jobq1_histogram} shows the histograms of the loss functions on the JQ1 dataset for \textbf{L} clustering. While the standard difference for the value in Eq. 4, is shown in Figure~\ref{figure:lda_jobq1_diff_boot} and Figure~\ref{figure:lda_jobq1_diff_cluster} for bootstrap and cluster samplers. In contrast to the jobs dataset (JQ1, JQ2, and JQ3), for NS and FE, \textbf{G} outperformed other models and it was followed by either \textbf{F} (for NS) or \textbf{K} (for FE). These results are related to the structure of the label distributions of the dataset. 

The hyperparameter for NBP is the neighborhood size ($r$). To build the synthetic dataset for NBP, we use the bootstrap and NBP based samplers defined in Algorithms \ref{alg:model_selection}--\ref{alg:model_selection_cluster}. Figure~\ref{figure:nbp_jobq1_diff_boot} and Figure~\ref{figure:nbp_jobq1_diff_nbp} shows the standard difference with linear piecewise fitting for bootstrap and NBP samplers. Table~\ref{table:nbp_model_selection} gives the optimum neighborhood sizes ($r$) based on each sampling method. Due to the nature of NBP, we also report $N_{Median}$, the median of all the neighborhood sizes. The $r$ sizes identified using the NBP sampler outperformed the sizes identified using the bootstrap sampler. In FE, both the values identified utilized the entire dataset which was available for pooling, while in contrast other datasets utilized approximately a quarter of the entire set. 
\begin{table}
\begin{center}
\caption{We achieve optimal label aggregation models on each label set with the presented number of clusters ($p$) and KL-divergence for the datasets using the cluster sampler with Multinational distribution as the loss function. \textit{The lowest KL per dataset is highlighted in blue.}}
\begin{tabular}{c|c|ccccc}
\textbf{Model} & &\textbf{JQ1} & \textbf{JQ2} & \textbf{JQ3} &\textbf{NS}&\textbf{FE} \\
\hline
\textbf{F}MM & $p$&\cellcolor{blue!25}29 &\cellcolor{blue!25} 12&\cellcolor{blue!25} 11 &36&16 \\
&$D_{KL}$&\cellcolor{blue!25}0.201& \cellcolor{blue!25}0.151 & \cellcolor{blue!25}0.347 &0.340&0.080\\
\hline
\textbf{G}MM &$p$ & 2 & 3 & 5 &\cellcolor{blue!25}6 &\cellcolor{blue!25}37\\
&$D_{KL}$&0.700 & 0.639 & 1.416 &\cellcolor{blue!25}0.215&\cellcolor{blue!25}0.008\\
\hline
\textbf{K}-Means&$p$ & 5 & 16 & 36 &6 &22\\
&$D{KL}$&0.716 & 0.809 &1.215 &0.654&0.049\\
\hline
\textbf{L}DA & $p$& 2    & 12        & 7  &4 &35 \\
&$D_{KL}$ & 0.428 & 0.201  & 0.587  &0.443&0.064\\
\hline
\end{tabular}
\label{table:model_selection_cluster_md}
\end{center}
\end{table}
\begin{table}
\begin{center}
\caption{We achieve optimal label aggregation models on each label set with the presented number of clusters ($p$) and KL-divergence for the datasets using the cluster sampler with KL-divergence as the loss function. \textit{The lowest KL per dataset is highlighted in blue.}}
\begin{tabular}{c|c|ccccc}
\textbf{Model} & &\textbf{JQ1} & \textbf{JQ2} & \textbf{JQ3} &\textbf{NS}&\textbf{FE} \\
\hline
\textbf{F}MM & $p$ &\cellcolor{blue!25}14      & \cellcolor{blue!25}7      & \cellcolor{blue!25}35 &7&16 \\
&$D_{KL}$&\cellcolor{blue!25}0.193      &\cellcolor{blue!25} 0.170     &\cellcolor{blue!25} 0.269 &0.935 &0.080\\
\hline
\textbf{G}MM &$p$ &  4 & 2 & 6 &\cellcolor{blue!25}17 &\cellcolor{blue!25}37\\
&$D_{KL}$&0.819 & 0.777 & 1.225 &\cellcolor{blue!25}0.285&\cellcolor{blue!25}0.008\\
\hline
\textbf{K}-Means&$p$ & 10 & 15 & 33 &12 &22\\
&$D{KL}$& 0.712 & 0.797 &1.126 &0.675&0.049\\
\hline
\textbf{L}DA & $p$& 7        & 5        & 5  &4 &35 \\
&$D_{KL}$ & 0.243       & 0.237  & 0.625  &0.602&0.064\\
\hline
\end{tabular}
\label{table:model_selection_cluster_kl}
\end{center}
\end{table}
\begin{table}
\center
\caption{We achieve optimal neighborhood size for NBP on each label set with the given KL-divergence and the median neighborhood size at each level. \textit{The lowest KL per dataset is highlighted in blue.}}

\begin{tabular}{c|ccccc}
& \textbf{JQ1} & \textbf{JQ2} & \textbf{JQ3} & \textbf{NS} &\textbf{FE} \\
\hline
$r_{B}$  & 5        & 6        & 8&4 &\cellcolor{blue!25}4.5 \\
$D_{KL}$ & 0.358     &  0.358    & 0.704&0.568 &\cellcolor{blue!25}0.080    \\
$N_{Median}$ & 967 &    922   & 863.5&548.5&\cellcolor{blue!25}1,250      \\
$N_{Maximum}$&1,000&1,000&1,000&1,000&\cellcolor{blue!25}1,250\\
\hline
$r_{NBP}$ &\cellcolor{blue!25} 3        &\cellcolor{blue!25} 2        & \cellcolor{blue!25}2      & \cellcolor{blue!25}2 &\cellcolor{blue!25} 3 \\
$D_{KL}$ &\cellcolor{blue!25} 0.317     &  \cellcolor{blue!25} 0.257   &\cellcolor{blue!25}0.409 &\cellcolor{blue!25}0.444 &\cellcolor{blue!25}0.080     \\
$N_{Median}$ &\cellcolor{blue!25} 875     &  \cellcolor{blue!25}645    &\cellcolor{blue!25}293 &\cellcolor{blue!25}265&\cellcolor{blue!25}1,250    \\
$N_{Maximum}$&\cellcolor{blue!25}1,000&\cellcolor{blue!25}1,000&\cellcolor{blue!25}1,000&\cellcolor{blue!25}1,000&\cellcolor{blue!25}1,250\\
\hline
\end{tabular}
\label{table:nbp_model_selection}
\end{table}

\subsection{Supervised Learning Experiments}
Following the same experimental process as Liu et al \cite{Liu2019HCOMP} we use the refined labels to train and test supervised learning models. Here, we tested our held-out data on two supervised learning models (CNN on Table~\ref{table:CNN_results} and LSTM on Table~\ref{table:LSTM_results}) from Liu et al., with the goal of predicting the pooled label distribution of each data item. 
Note that the features for the supervised learning models to learn were different among the datasets. The JQ1, JQ2, and JQ3 datasets are text-based \cite{Liu2019HCOMP} and the NS and FE  features are the vectorized representations used by Geng \cite{Geng2015}. 
We evaluated the predictions against the refined labels generated from the clustering and NBP. We evaluated them: (1) as a traditional supervised learning problem by measuring the accuracy (using $argmax(Y_{K_{k(x)}})$), (2) as a probability distribution problem using the KL-divergence. 
 
 \subsubsection{Results from Label Clustering}
Looking at the KL-divergence results, the CNNs in Table~~\ref{table:CNN_results} outperformed the LSTM models. In all the instances, the labels refined through clustering outperformed the baseline. The labels refined using the \textbf{G} models outperformed the other models for the JQ1 and JQ2 datasets while for JQ3 and NS, \textbf{K} outperformed the others. For the FE dataset, the labels refined by the \textbf{F} model performed better than the other models. These results could be related to the size of the label spaces. JQ3 had the largest label space, while others were within a similar range. Moreover, in JQ3, human annotators were allowed to provide multiple labels per data item, in contrast to others where only one label choice was allowed per item. In the LSTMs, the observations from the CNNs were common, in contrast to JQ1, where \textbf{L} outperformed other models.  

 \subsubsection{Results from Label Pooling}
In our experiments, we used the two radii picked using bootstrap and NBP sampling. Looking at the mean KL-divergence, for the CNNs, the NBP outperformed the clustering models. The accuracy results obtained did supplement the results obtained through the KL test in majority of the cases other than FE dataset. Similar to the clustering approaches, the results from CNNs outperformed the results obtained through the LSTMs. These observations will be discussed further in the next section.

\begin{table*}[t]
\centering
\label{tab:CNN_results}
\caption{The predictions with KL-divergence ($\mathbf{D_{KL}}$) when used with supervised learning (CNN) and unsupervised learning (clustering and NBP). The \textit{lowest} KL and the \textit{highest} accuracy for NBP and clustering is highlighted in blue.}

\begin{tabular}{c|c|cccc|cc||c|cccc|cc}
\multicolumn{8}{c}{KL-divergence} & \multicolumn{7}{c}{Accuracy}\\ 
\textbf{Dataset}    & \textbf{Raw} & \multicolumn{4}{c|}{\textbf{Clustering}} & \multicolumn{2}{c||}{\textbf{NBP} - KL} &\textbf{Raw} & \multicolumn{4}{c|}{\textbf{Clustering}} & \multicolumn{2}{c}{\textbf{NBP} - KL}       \\
& \textbf{Labels} & \textbf{F}  & \textbf{G} & \textbf{K} & \textbf{L}& $r_{B}$  &  $r_{NBP}$ & \textbf{Labels} & \textbf{F}  & \textbf{G} & \textbf{K} & \textbf{L}& $r_{B}$  &  $r_{NBP}$   \\
        \hline
\multicolumn{15}{c}{Jobs dataset - Supervised Learning Classification (CNN)}\\ 
        \hline
JQ1 & 1.088 & 0.346  & \cellcolor{blue!25}0.265 &0.270 &0.370 &\cellcolor{blue!25}0.021 & 0.028 & 0.537&\cellcolor{blue!25}0.747 & 0.575  &0.677 &0.727            & \cellcolor{blue!25}1.000 & 0.974 \\
JQ2 &1.072& 0.483    & \cellcolor{blue!25}0.148 &0.306 &0.738&0.064 &\cellcolor{blue!25}0.032 &0.477 & 0.720 & \cellcolor{blue!25} 1.000   &0.652 &0.648          & 0.916 & \cellcolor{blue!25}1.000 \\
JQ3 &1.440& 0.772  &  0.366 &\cellcolor{blue!25}0.341 &1.033& 0.206 &\cellcolor{blue!25}0.117 &0.271 & 0.313    & 0.467   &\cellcolor{blue!25}0.553 &0.330       & 0.528 &\cellcolor{blue!25} 0.600   \\
\hline
\multicolumn{15}{c}{Natural Scenes dataset - Supervised Learning Classification (CNN)}\\ 
\hline
NS & 0.985 & 0.589  & 0.469 &\cellcolor{blue!25}0.235 & 0.551 &\cellcolor{blue!25} 0.117 &0.192 &0.282 & 0.816  & 0.186 &0.340 & \cellcolor{blue!25}0.828 & \cellcolor{blue!25}0.529 & 0.044\\
\hline
\multicolumn{15}{c}{Facial Expressions dataset - Supervised Learning Classification (CNN)}\\ 
\hline
FE & 0.081 &\cellcolor{blue!25}0.0009  & 0.071 &0.081 & 0.081 & 2.927 &\cellcolor{blue!25}1.045 &0.259 & 0.353  & 0.315 &0.259 & \cellcolor{blue!25}0.227 & \cellcolor{blue!25}1.000 & 1.000\\
\hline

\end{tabular}
\label{table:CNN_results}
\end{table*}
\begin{table*}[h]
\centering
\caption{The predictions with KL-divergence ($\mathbf{D_{KL}}$) when used with supervised learning (LSTM) and unsupervised learning (clustering and NBP). The \textit{lowest} KL and the \textit{highest} accuracy for NBP and clustering is highlighted in blue.}
\begin{tabular}{c|c|cccc|cc||c|cccc|cc}
\multicolumn{8}{c}{KL-divergence} & \multicolumn{7}{c}{Accuracy}\\ 
\textbf{Dataset}    & \textbf{Raw} & \multicolumn{4}{c|}{\textbf{Clustering}} & \multicolumn{2}{c||}{\textbf{NBP} - KL} &\textbf{Raw} & \multicolumn{4}{c|}{\textbf{Clustering}} & \multicolumn{2}{c}{\textbf{NBP} - KL}       \\
& \textbf{Labels} & \textbf{F}  & \textbf{G} & \textbf{K} & \textbf{L}& $r_{B}$  &  $r_{NBP}$ & \textbf{Labels} & \textbf{F}  & \textbf{G} & \textbf{K} & \textbf{L}& $r_{B}$  &  $r_{NBP}$   \\
        \hline
\multicolumn{15}{c}{Jobs dataset - Supervised Learning Classification (LSTM)}\\ 
        \hline
JQ1 & 0.907 & 0.874 & 0.918 & 0.788 &\cellcolor{blue!25} 0.770  & 0.550 & \cellcolor{blue!25}0.533& 0.865 & \cellcolor{blue!25}0.874 & 0.794 & 0.861 & 0.844 & 0.987 & \cellcolor{blue!25} 1.000  \\
JQ2 & 0.989 &1.038 & \cellcolor{blue!25}0.624 & 0.746 & 1.084& \cellcolor{blue!25}0.659 & 0.669& 0.841 & 0.861 &\cellcolor{blue!25} 1.000 & 0.855 & 0.832  & 0.964 & \cellcolor{blue!25}1.000    \\
JQ3 & 1.567 &1.358 & 1.003 & \cellcolor{blue!25}0.910 & 1.456& \cellcolor{blue!25}0.789 & 0.748& 0.643 & 0.621 & 0.642 & \cellcolor{blue!25}0.718 & 0.612    & \cellcolor{blue!25}0.782 & 0.821    \\

\hline
\end{tabular}
\label{table:LSTM_results}
\end{table*}

\section{Discussion}
Discrepancies in the results obtained from different clustering models correspond to discrepancies in the label distributions in different datasets and in the behaviors of different clustering models. For instance, the label distributions in the jobs dataset (Figure \ref{Fig:Q1_LD}) tend to be skewed towards one label choice, while this was not the case for NS (Figure \ref{Fig:NS_LD}) and FE (Figure \ref{Fig:FE_LD}). This skewing in the jobs dataset label distributions may be due to the nature of human annotation with respect to human interpretations of the questions asked about the data. 

In contrast to the clustering methods, NBP as a whole showed promising results throughout the experiments. On the NS and FE datasets, it showed a pattern of increasing KL divergence as the radius increased. This behavior seems to depend on how the label distributions are structured. For instance, the skewed label distributions in the jobs dataset results in some items having no neighbors at smaller radii. 


The jobs dataset by Liu et al. \cite{Liu2019HCOMP} contains labels obtained from from two crowdsourcing platforms. Manual inspection suggests there exist population-level disparities in the distributions provided by each. This phenomenon could be studied more rigorously by modeling user behaviors and traits to improve PLDL. One approach, for instance, could involve inter-annotator agreement using measures such as Cohen's Kappa \cite{LAZAR2017299} to either resolve conflicts during annotation or use them for refining the distribution estimates. However, a challenge would be to acquire a dataset that contains annotator information, as they are generally removed in publicly available datasets.  

Table~\ref{table:nbp_model_selection} raises questions about hyperparameter selection for NBP. As our main goal is to share labels between neighbors, looking at the median number of neighbors per each data item ($N_{Median}$) presents some insights. All the datasets had a $N_{Median}$ of at least a quarter of the entire dataset except for FE. One reason could be that FE is, among the datasets we consider, the one that has the largest population of annotators. 

While single label learning has a number of established performance measures, such measures are not so well-established in label distribution learning.  Geng \cite{Geng2015} analyzed 41 different measures and identified five (KL-divergence, Chebyshev, Clark, Canberra, Cosine similarity, and Intersection) as most effective. In our work, we used KL-divergence (one of the five measures identified). KL-divergence is used to measure information loss specifically for probability distributions. The use of our sampling techniques to evaluate the models and hyperparameter selection contributes to the establishment of standard procedures for LDL evaluation. 

\section{Conclusion}
Gathering labeled data is an evident resource bottleneck for population-based label distribution learning (PLDL). We introduced and studied new methods for refining the label distributions of data items for PLDL based on the labels of their neighbors. Neighborhood-based pooling (NBP) is semi-supervised learning approach that uses information theoretic measures to pool similar label distributions. We compared NBP as a pooling technique for supervised learning to clustering methods introduced in prior research. We also introduced new methods based on population hypothesis testing for selecting models for label refinement. Our results show that NBP is a feasible approach for refining label distribution estimates. 

\section{Acknowledgments}
We thank the anonymous reviewers for their helpful feedback and suggestions. Many thanks to Ifeoma Nwogu, Cecilia O. Alm, Victoria Maung, James Spann, and Nandula Perera for their contributions and conversations.
\bibliography{1366_paper}
\appendix
\section{Appendix}
\begin{figure}[h]
\begin{minipage}[t]{0.3\linewidth}
    \includegraphics[width=\linewidth]{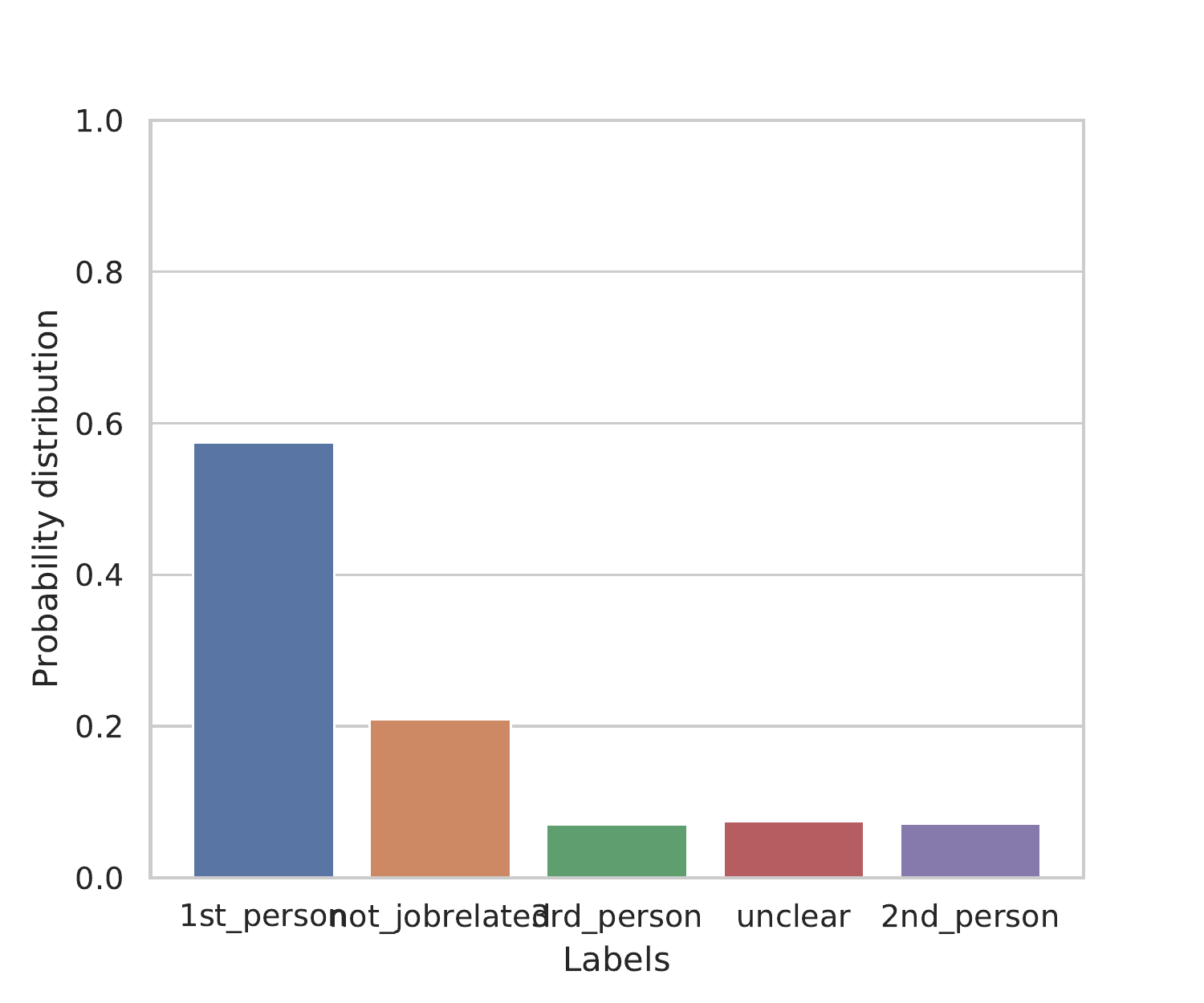}
  \caption{Overall label distribution of the JQ1 dataset \cite{Liu2019HCOMP}.}
 \label{Fig:Q1_LD}
\end{minipage}%
    \hfill%
\begin{minipage}[t]{0.3\linewidth}
    \includegraphics[width=\linewidth]{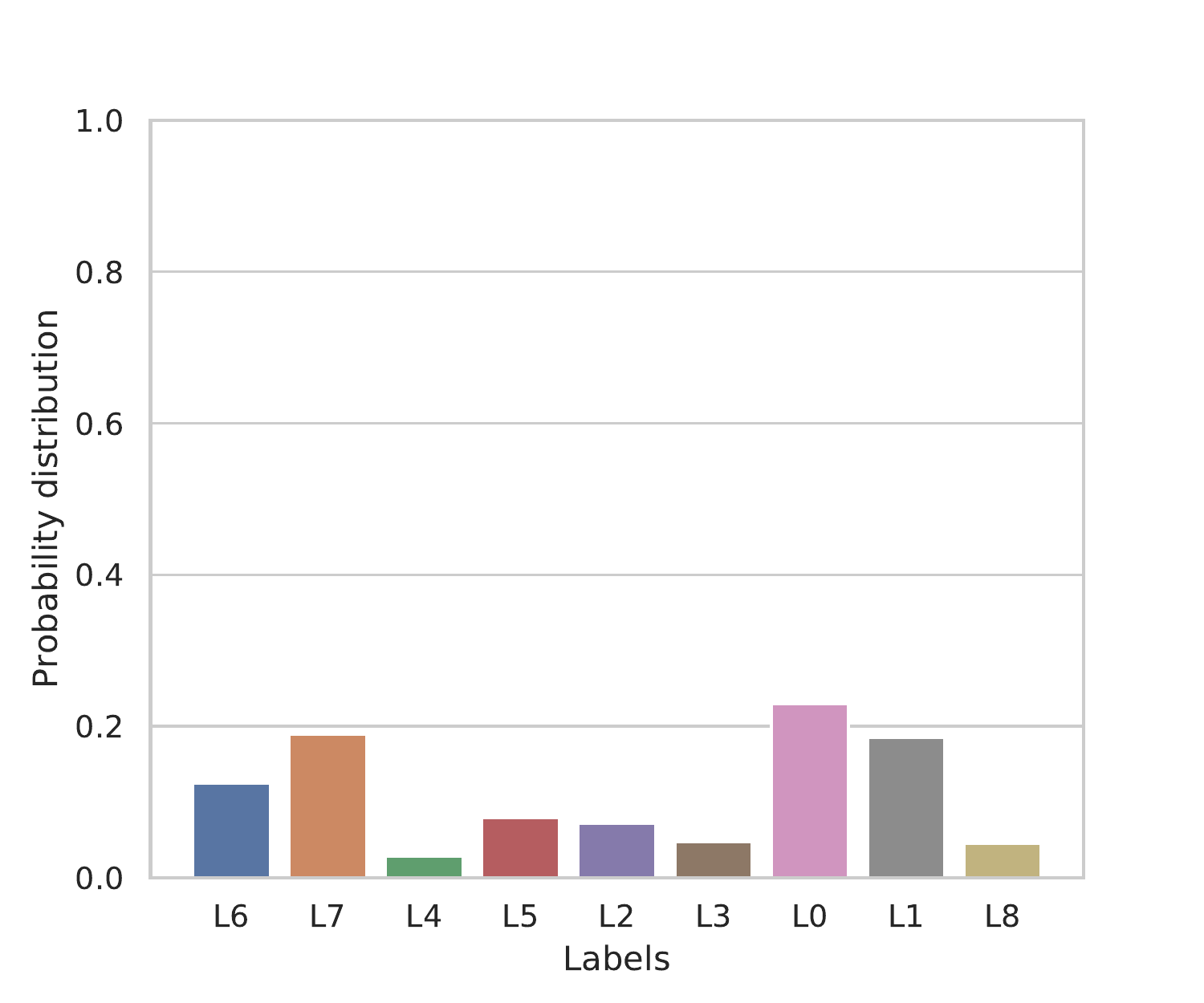}
  \caption{Overall label distribution of the NS dataset \cite{Geng2015}.}
 \label{Fig:NS_LD}
\end{minipage} 
    \hfill%
\begin{minipage}[t]{0.3\linewidth}
    \includegraphics[width=\linewidth]{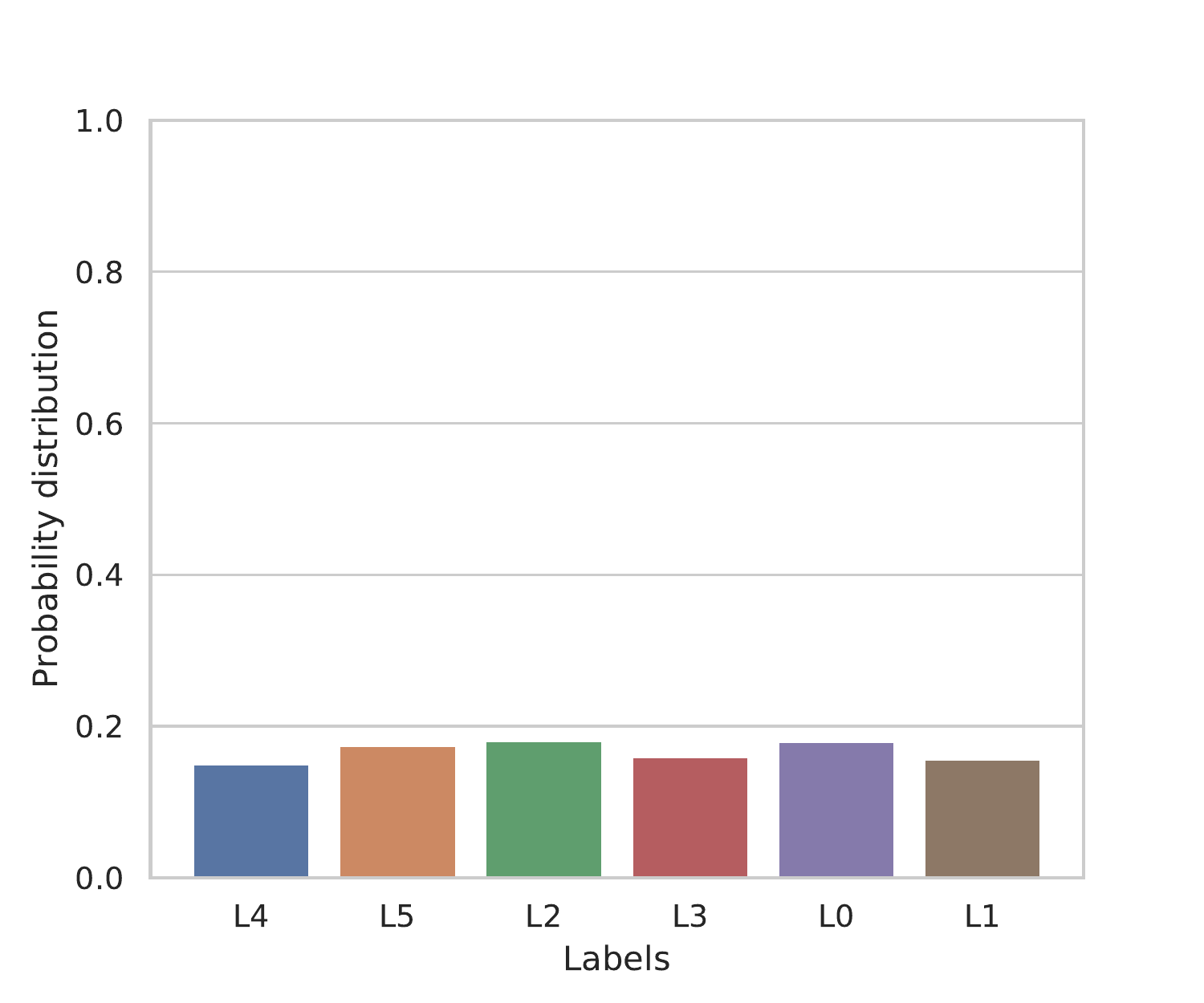}
  \caption{Overall label distribution of the FE dataset \cite{face_dataset}.}
 \label{Fig:FE_LD}
\end{minipage} 
\end{figure}
\end{document}